\documentclass[conference,a4paper]{IEEEtran}
\IEEEoverridecommandlockouts

\usepackage[hidelinks]{hyperref}
\usepackage[cmex10]{amsmath}
\usepackage{amssymb,amsfonts}
\usepackage{dblfloatfix}

\usepackage[ruled,vlined]{algorithm2e}
\usepackage{graphicx}
\graphicspath{{Figures/PDF/}{Figures/PNG/}}
\usepackage{multirow}

\usepackage{booktabs}
\usepackage{siunitx}
\usepackage[numbers,compress]{natbib}
\usepackage{texnames}
\usepackage{bm,bbm}
\usepackage{orcidlink}

\begin{document}

\title{\uppercase{Joint Retrieval of Cloud  properties using Attention-based Deep Learning Models}

\thanks{Copyright 2024 IEEE. Published in the 2025 IEEE International Geoscience and Remote Sensing Symposium (IGARSS 2025), scheduled for 3 - 8 August 2025 in Brisbane, Australia. Personal use of this material is permitted. However, permission to reprint/republish this material for advertising or promotional purposes or for creating new collective works for resale or redistribution to servers or lists, or to reuse any copyrighted component of this work in other works, must be obtained from the IEEE. Contact: Manager, Copyrights and Permissions / IEEE Service Center / 445 Hoes Lane / P.O. Box 1331 / Piscataway, NJ 08855-1331, USA. Telephone: + Intl. 908-562-3966.}
}

\author{	
    \IEEEauthorblockN{Zahid Hassan Tushar\orcidlink{0000-0002-8231-6767}, Adeleke Ademakinwa, Jianwu Wang, Zhibo Zhang, and Sanjay Purushotham}
    \IEEEauthorblockA{\textit{University of Maryland, Baltimore County, Maryland, USA.}\\
        Emails: \{ztushar1, adeleka1, jianwu, zzbatmos, psanjay\}@umbc.edu}
}

\maketitle
\begin{abstract}

Accurate cloud property retrieval is vital for understanding cloud behavior and its impact on climate, including applications in weather forecasting, climate modeling, and estimating Earth's radiation balance. The Independent Pixel Approximation (IPA), a widely used physics-based approach, simplifies radiative transfer calculations by assuming each pixel is independent of its neighbors. While computationally efficient, IPA has significant limitations, such as inaccuracies from 3D radiative effects, errors at cloud edges, and ineffectiveness for overlapping or heterogeneous cloud fields.
Recent AI/ML-based deep learning models have improved retrieval accuracy by leveraging spatial relationships across pixels. However, these models are often memory-intensive, retrieve only a single cloud property, or struggle with joint property retrievals. To overcome these challenges, we introduce CloudUNet with Attention Module (CAM), a compact UNet-based model that employs attention mechanisms to reduce errors in thick, overlapping cloud regions and a specialized loss function for joint retrieval of Cloud Optical Thickness (COT) and Cloud Effective Radius (CER).
Experiments on a Large Eddy Simulation (LES) dataset show that our CAM model outperforms state-of-the-art deep learning methods, reducing mean absolute errors (MAE) by 34\% for COT and 42\% for CER, and achieving 76\% and 86\% lower MAE for COT and CER retrievals compared to the IPA method.
\end{abstract}


\section{Introduction}
Clouds are a key regulator of Earth's radiation balance, controlling how much radiation is absorbed or reflected back into space. Their microphysical properties, such as Cloud Optical Thickness (COT) and Cloud Effective Radius (CER), are essential for climate modeling and weather forecasting, providing critical insights into climate change~\cite{nakajima1990determination, ipcc_website, ma2019retrieval}.
Retrieving cloud properties from satellite radiance observations is a 3D inverse problem because radiance observations are collected from 3D clouds and are affected by different cloud properties. The complex interactions between clouds and solar radiation, described by 3D radiative transfer models, require a 3D inversion process to retrieve the underlying cloud properties. However, since radiance observations are inherently 2D, they lack complete information about cloud and solar interactions, making the inversion mathematically intractable and necessitating approximate solutions.
The Independent Pixel Approximation (IPA), proposed by~\citet{nakajima1990determination}, offers a 1D inverse method for cloud property retrieval. While computationally efficient, IPA often leads to overestimations or underestimations~\cite{li2013new, baum2012modis, tushar2024cloudunet}.

In recent years, machine learning and deep learning-based techniques have been adopted in many fields, including computer vision~\cite{rombach2022high}, medical applications~\cite{rahman2023communication}, robotics~\cite{hossain2023covernav}, and edge devices~\cite{anwar2023heteroedge}, as well as in remote sensing~\cite{mostafa2023cnn,mostafa2025gwavenet}. In particular, their application to cloud property retrieval has significantly narrowed the gap between true and retrieved values.~\cite{li2024physics}. For example,~\citet{min2020retrieval} explored four ML algorithms such as K-Nearest-Neighbors (KNN)\cite{altman1992introduction}, Support Vector Machines (SVM) \cite{drucker1996support}, Random Forests (RF) \cite{breiman2001random}, and Gradient Boosting Decision Trees (GBDT) \cite{friedman2002stochastic} to retrieve Cloud Top Height (CTH) from radiance observations. Similarly, \citet{yang2022machine}  trained the XGBoost algorithm to retrieve CTH, cloud masks, and cloud top temperature from Himawari-8 measurements~\cite{letu2018ice}. However, these approaches neglect spatial features, limiting their performance~\cite{nataraja2022segmentation, tushar2024cloudunet}. 
Recent efforts have focused on utilizing spatial regions to retrieve cloud properties, moving beyond pixel-by-pixel approaches. These methods fall into two categories: single-property retrievals and joint-property retrievals. Single-property methods often train separate UNet-style architectures~\cite{li2019residual, 9553646} for each cloud property, such as COT, CER, and CTH~\cite{nataraja2022segmentation, li2023transfer, zhao2024image, tushar2024cloudunet}. However, maintaining separate models for each property incurs significant computational costs.
To address this, \citet{okamura2017feasibility} proposed a single DNN for joint retrieval of COT and CER. Later, \citet{wang2022cloud} introduced a more advanced DNN architecture to jointly retrieve cloud masks, CTH, and ice Cloud Optical Thickness using temperature brightness and auxiliary information. However, their use of a large spatial region $(32\times32)$ to retrieve a single pixel’s properties makes their approach computationally intensive during inference. Similarly, \citet{wang2022retrieval} trained a UNet-based model to jointly estimate COT, CER, CTH, and cloud masks. Despite these advancements, these methods are either computationally expensive due to the requirement for high-dimensional radiance data or produce less accurate retrievals.
In this paper, we address the limitations of existing methods by proposing an efficient deep learning model with a custom objective function to effectively learn the joint distribution of COT and CER from radiance data while mitigating 3D radiative effects. Our key contribution is the development of CloudUNet with Attention Module (CAM), a lightweight UNet-based architecture that uses attention mechanisms to reduce retrieval errors in dense, overlapping cloud regions and incorporates a multi-task loss function for joint COT and CER retrieval. The experimental results show that CAM achieves at least 34\% lower MAE compared to state-of-the-art deep learning methods and 76\% lower MAE compared to the IPA method. Ablation studies further demonstrate the effectiveness of the attention mechanism and the custom objective function in reducing retrieval errors.

\section{Methodology}
\subsection{Problem Formulation}
We formulate the joint retrieval of cloud properties (such as COT and CER) from radiance observations as a multi-task regression problem. Mathematically, the joint retrieval formulation for a cloud profile ($\mathcal{C}$) is shown in Eq.~\ref{eq.1}
\begin{equation}\label{eq.1}
\tau^{\mathcal{C}}, r^{\mathcal{C}}_e = f_{\theta}(R^{\mathcal{C}}_{0.66},R^{\mathcal{C}}_{2.13})
\end{equation}
where $\tau^{\mathcal{C}}$ is the COT,
$r^{\mathcal{C}}_e$ is the CER,
$f_{\theta}$ is the model parameters, and
$R^{\mathcal{C}}_{0.66}, R^{\mathcal{C}}_{2.13}$ are the radiance observations at wavelengths $0.66\mu m$ and $2.13\mu m$.
\subsection{Cloud Property Retrieval framework}
We employ a window-based retrieval approach to estimate the cloud properties from the radiance observations. In this approach, a small spatial segment of the radiance measurements is extracted for each cloud profile and input into the deep learning algorithm to estimate cloud properties for that specific segment. A hash map is used to traverse the radiance observations, extract patches, and generate the corresponding cloud property estimates. Subsequently, this hash map is then used to aggregate the properties from all segments to obtain the full cloud profile properties.

Window-based retrievals offer several advantages, such as enabling the reuse of deep learning models for larger radiance observations and reducing the need for extensive training data due to the small spatial resolution of the extracted patches. This approach also enhances the model's generalization to unseen test data. Researchers have explored various window sizes for retrieving cloud properties from satellite radiance observations. In this work, we adopt the most commonly used window size of $64 \times 64$~\cite{nataraja2022segmentation,okamura2017feasibility,li2023transfer}. 

\subsection{CAM: CloudUNet with Attention Module}

In our previous work~\cite{tushar2024cloudunet} presented at IGARSS'24, we introduced CloudUNet, a state-of-the-art deep learning model for COT retrieval. CloudUNet employs a compact U-Net-style architecture with skip connections to learn multi-scale features, significantly reducing errors in cloud property retrieval. However, despite its superior performance, CloudUNet has limitations: its accuracy decreases when handling heterogeneous clouds, and it is designed to retrieve only one cloud property at a time.
To address these challenges, this paper introduces CAM: CloudUNet with Attention Module, which incorporates channel attention modules~\cite{woo2018cbam} into the skip connections. These modules enhance the model's ability to focus on the most relevant cloud patterns in radiance observations, improving the accuracy of cloud property predictions. Additionally, CAM is designed to jointly predict both COT and CER using two separate branches with convolutional layers, as illustrated in Figure 1.

Intuitively, CAM works as follow: It extracts features from the radiance data at multiple scales using two downsampling blocks and several convolutional layers. These features are then mapped back to the target COT and CER data through two upsampling blocks and additional convolutional layers. The \textit{Channel Attention Module} refines the features by re-weighting them along the channel dimensions, assigning greater weight to important features while reducing the weight of less significant ones. Additional branches for each cloud property leverage features from earlier layers to map to the corresponding target cloud property.

\begin{figure}
  \centering
  \includegraphics[width=\linewidth]{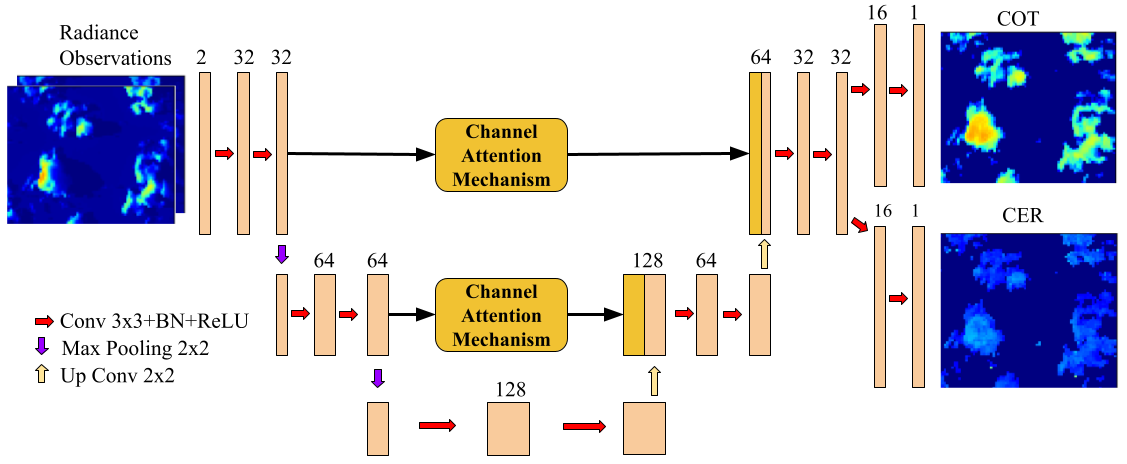}
  \caption{CAM: CloudUNet with Attention Module}
  \label{fig:cam}
  \vspace{-10pt}
\end{figure}

\subsection{Training Objective Function}
In regression-based problems, it is common practice to use L2 loss for training deep learning models. However, different cloud properties have distinct value ranges, and L2 loss imposes disproportionately larger penalties for higher values while assigning smaller penalties to lower values. This approach fails to account for disparities between cloud properties with narrow value ranges (e.g., scaled CER) and those with broader value ranges (e.g., COT). As a result, balancing the loss contributions from different properties is essential. To address this, we introduce a multi-task objective (MTO) loss that incorporates two weighting coefficients, $\lambda_1$ and $\lambda_2$,  which are applied to the L2 losses for COT and CER, respectively. The L2 loss formula is provided in Eq.\ref{eq.2}, and the MTO loss formulation ($L_{MTO}$) used in our training is detailed in Eq.\ref{eq.3}.

\begin{equation}
    l(t,y)=\frac{1}{N}\sum_{i=1}^{N}(t_i-y_i)^2
    \label{eq.2}
\end{equation}

\begin{equation}
    L_{MTO} = \lambda_1 \cdot l_{COT}+ \lambda_2 \cdot l_{CER}
    \label{eq.3}
\end{equation}

Here, $t$ and $y$ are the true and the retrieved cloud property respectively. $i$ denotes $i^{th}$ pixel, and $N$ denotes the total pixels in a cloud profile. $l_{COT}$ and $l_{CER}$ denote L2 losses for COT and CER respectively.

\begin{table}[ht]
    \centering
    \caption{Comparison of COT and CER Retrievals Methods. COT Scores.}
    \label{tab:results_p1}

    \resizebox{\columnwidth}{!}{%
    \begin{tabular}{r c c c c}
        \toprule
\textbf{\begin{tabular}[c]{@{}c@{}}Retrieval \\ Methods\end{tabular}} & \textbf{\begin{tabular}[c]{@{}c@{}}COT\\ MAE($\downarrow$) \end{tabular}} & \textbf{\begin{tabular}[c]{@{}c@{}}COT \\ MSE ($\downarrow$)\end{tabular}} & \textbf{\begin{tabular}[c]{@{}c@{}}COT \\ Corr. Coeff. ($\uparrow$)\end{tabular}} & \textbf{\begin{tabular}[c]{@{}c@{}} \% Impv. Over\\IPA MAE\end{tabular}} \\
        \midrule
        IPA Retrievals~\cite{nakajima1990determination}  & 
        $0.176 \pm 0.003$ & $0.199 \pm 0.005$ & $0.918 \pm 0.002$ & -- \\
        
        Okamura~\cite{okamura2017feasibility}  & 
        $0.093 \pm 0.001$ & $0.030 \pm 0.001$ & $0.985 \pm 0.001$ & $47\%$ \\
        
        UNet~\cite{nataraja2022segmentation,zhao2023cloud}  & 
        $0.065 \pm 0.008$ & $0.019 \pm 0.001$ & $0.991 \pm 0.001$ & $63\%$ \\
        
        CloudUNet~\cite{tushar2024cloudunet}  & 
        $0.070 \pm 0.007$ & $0.019 \pm 0.002$ & $0.991 \pm 0.001$ & $60\%$ \\
        
        \textbf{CAM (ours)}  & 
        $\mathbf{0.043} \pm \mathbf{0.001}$ & 
        $\mathbf{0.014} \pm \mathbf{0.001}$ & 
        $\mathbf{0.993} \pm \mathbf{0.0003}$ & 
        $\mathbf{76\%}$ \\
        \bottomrule
    \end{tabular}
    }
\end{table}

\begin{table}[ht]
    \centering
    \caption{Comparison of COT and CER Retrievals Methods. CER Scores.}
    \label{tab:results_p2}
    \resizebox{\columnwidth}{!}{%
    \begin{tabular}{r c c c c}
        \toprule
\textbf{\begin{tabular}[c]{@{}c@{}}Retrieval \\ Methods\end{tabular}} & \textbf{\begin{tabular}[c]{@{}c@{}}CER\\MAE($\downarrow$) \end{tabular}} & \textbf{\begin{tabular}[c]{@{}c@{}}CER\\MSE ($\downarrow$)\end{tabular}} & \textbf{\begin{tabular}[c]{@{}c@{}}CER \\ Corr. Coeff.($\uparrow$)\end{tabular}} & \textbf{\begin{tabular}[c]{@{}c@{}} \% Impv. Over\\IPA MAE\end{tabular}} \\
        \midrule
        IPA Retrievals~\cite{nakajima1990determination} & 
        $2.024 \pm 0.045$ & $34.536 \pm 1.291$ & $0.598 \pm 0.003$ & -- \\
        
        Okamura~\cite{okamura2017feasibility} & 
        $0.814 \pm 0.029$ & $2.042 \pm 0.084$ & $0.876 \pm 0.006$ & $60\%$ \\
        
        UNet~\cite{nataraja2022segmentation,zhao2023cloud} & 
        $0.435 \pm 0.092$ & $0.895 \pm 0.079$ & $0.947 \pm 0.003$ & $78\%$ \\
        
        CloudUNet~\cite{tushar2024cloudunet} & 
        $0.407 \pm 0.049$ & $0.919 \pm 0.117$ & $0.946 \pm 0.002$ & $80\%$ \\
        
        \textbf{CAM (ours)} & 
        $\mathbf{0.252} \pm \mathbf{0.010}$ & 
        $\mathbf{0.642} \pm \mathbf{0.033}$ & 
        $\mathbf{0.961} \pm \mathbf{0.004}$ & 
        $\mathbf{86\%}$ \\
        \bottomrule
    \end{tabular}
    }
\end{table}

\section{Experimental Setup}

\subsection{Data}

In this study, 902 Large-Eddy Simulation (LES) cloud fields (denoted as cloud profile) have been modeled using three-dimensional (3D) cloud Liquid Water Content (LWC) obtained from the LES Atmospheric Radiation Measurement (ARM) Symbiotic Simulation and Observation (LASSO)\cite{gustafson2020large}. The cloud droplet effective radius associated with the corresponding LWC distribution was calculated using a two-moment bulk microphysics scheme~\cite{morrison2008new} [see their equation 5 in Section 2]. The radiative transfer calculation was conducted using a solar zenith angle (SZA) of $60^\circ$, a solar azimuth angle (SAA) of $0^\circ$, and a view zenith angle (VZA) of $0^\circ$, with double periodic horizontal boundary conditions applied.
The surface was modeled as a Lambertian reflector with an albedo of $0.05$. Radiance computations (3D) at the two wavelengths ($0.66$ and $2.13 \mu$m), which are required for the \cite{nakajima1990determination} IPA bi-spectral retrievals, have been simulated using the spherical harmonics discrete ordinate method (SHDOM) developed by Evans et al.\cite{evans1998spherical}.

 The LES cloud fields from LASSO were selected because they provide a realistic representation of the atmosphere and cloud, consistent with observations from the ARM program. These cloud fields have a large domain size of $14 km \times 14 km$ with a horizontal resolution of $100m$. The vertical domain height extends up to $15km$, featuring a vertical resolution of $30m$ below $5km$ and $300m$ above $5km$. The large domain size and high resolution of the LASSO LES cloud fields make them suitable to study 3D radiative transfer effects in cloud properties retrievals from radiance observations. Each cloud profile has COT$[Dim: 144\times144\times1]$, CER$[Dim: 144\times144\times1]$ and Radiance observations at two wavelengths ($0.66 \mu m, 2.13 \mu m$) $[Dim: 144\times144\times2]$. Our goal is to retrieve the COT and CER from the radiance observations.

\textit{Data Preparation:} The  COT data are transformed using a shifted logarithmic transformation [$\text{Transformed COT} = log(COT + 1)$] to handle its sparse tail distribution\cite{tushar2024cloudunet}. Additionally, CER data are scaled by $30.0$ to be within a small range of $[0,2]$. This scaling is beneficial since it provides numerical stability for deep learning training\cite{rombach2022high}.

\subsection{Model Comparison}

We compared the following retrieval methods: (a) IPA retrievals by~\cite{nakajima1990determination}, a physics-based method used by NOAA; (b) DNN~\cite{okamura2017feasibility}; (c) UNet~\cite{nataraja2022segmentation, zhao2023cloud}; (d) CloudUNet~\cite{tushar2024cloudunet}; and (e) our CAM model. To evaluate performance, we employed three metrics: mean squared error (MSE)\cite{goodfellow2016deep}, mean absolute error (MAE)\cite{willmott2005advantages}, and the Pearson correlation coefficient~\cite{pearson1896vii}. These metrics offer insights into the magnitude of errors, the presence of bias in retrievals, and the linear correlation between true and retrieved properties.

\subsection{Implementation}
All methods are implemented in python using Pytorch library package and trained with various hyperparameters such as learning rate, batch size, schedulers, and optimizers. Our CAM model achieved the best performance with 0.001 learning rate, 128 batch size, and an early stopping criterion and patience of 50 epochs. The model parameters were optimized using the ADAM optimizer, with weighting coefficients $\lambda_1=1$ \& $\lambda_2=15$.

\footnotemark[1]
\footnotetext[1]{Our code is available at: \url{https://github.com/AI-4-atmosphere-remote-sensing/DL_3d_cloud_retrieval/releases}.}

\section{Results and Discussion}
\subsection{Quantitative Results}

Tables~\ref{tab:results_p1} and \ref{tab:results_p2} summarize the performance of all models. Our CAM outperformed all other methods by a significant margin, achieving at least 34\% lower MAE for COT and 28\% lower MAE for CER compared to state-of-the-art models. Additionally, Pearson correlation coefficient analysis reveals that the CAM model achieves the highest correlation. These improvements can be attributed to the use of multiscale features with the channel attention mechanism and further enhanced by a customized objective function, which is effectively utilized through separate cloud property branches [see section~\ref{sec:ablation}].

\subsection{Qualitative Results}

\begin{figure}
  \centering
  \includegraphics[width=\linewidth]{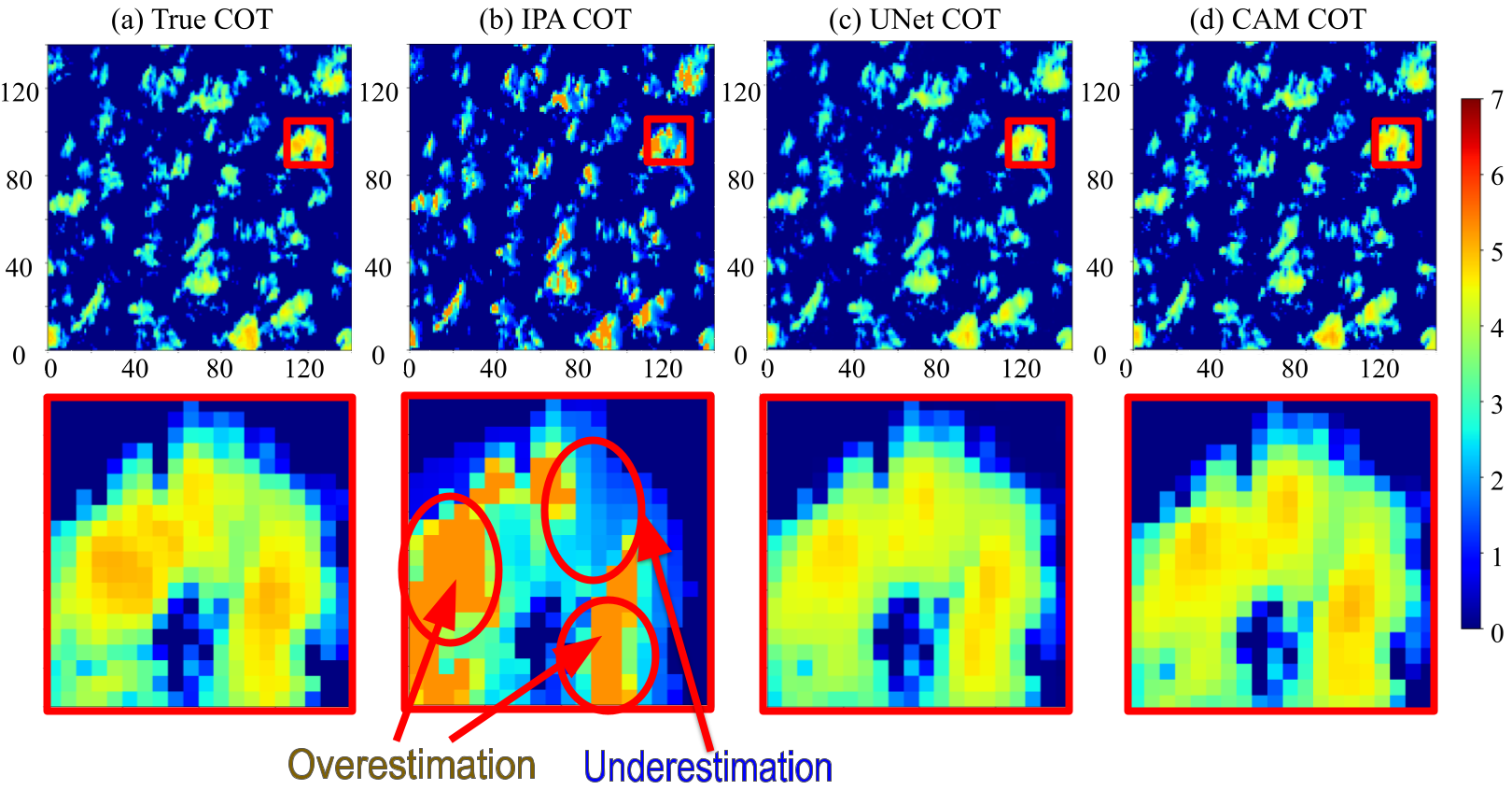}
  \caption{Comparison of COT Retrieval methods. COT values are plotted in shifted log scale. (a) True COT; (b) IPA retrieved COT; (c) UNet retrieved COT; (d) CAM retrieved COT. Highlighted regions are shown in bottom row.}
  \label{fig:resultscot}
\end{figure}

\begin{figure}
  \centering
  \includegraphics[width=\linewidth]{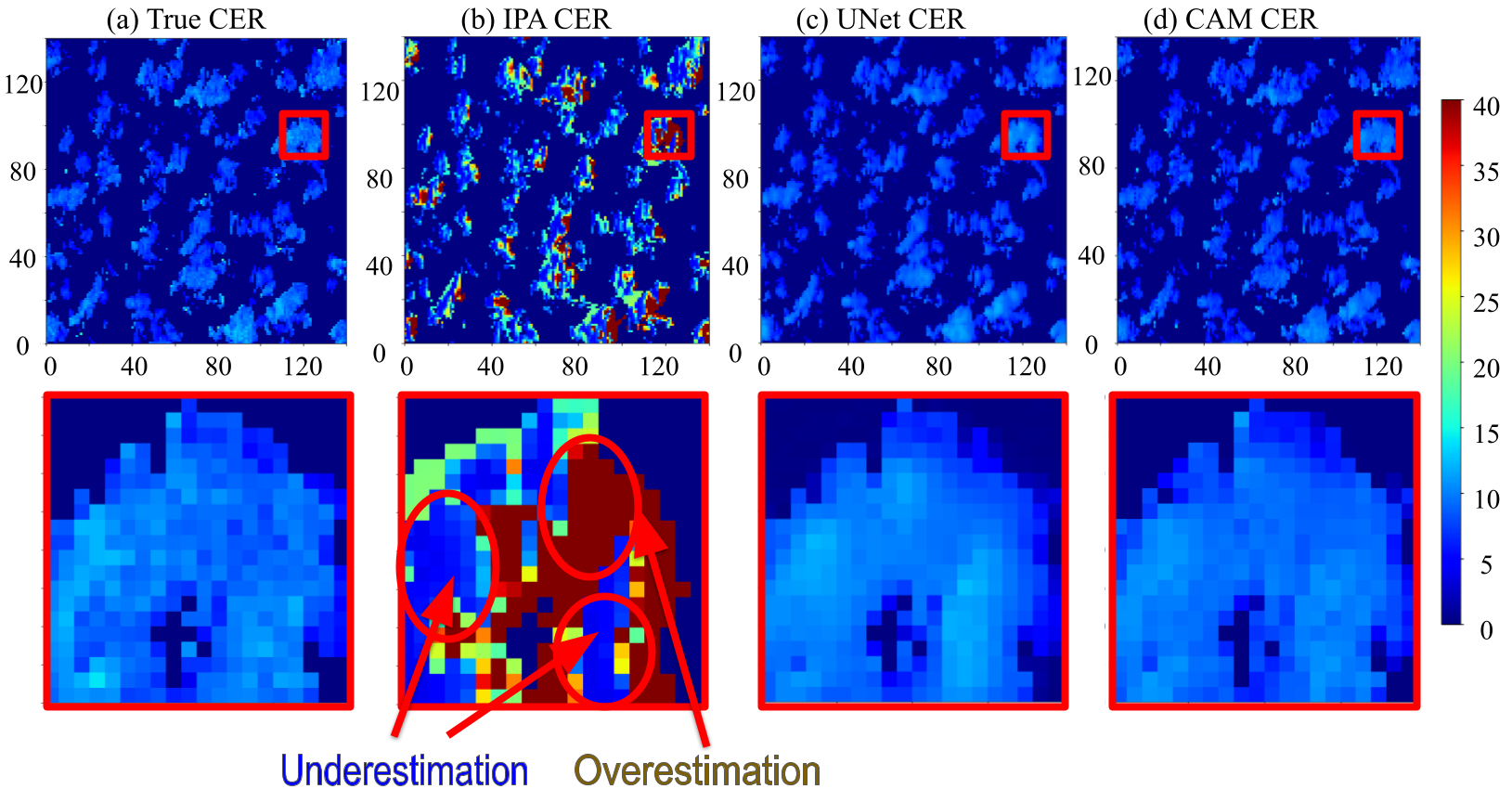}
  \caption{Comparison of CER Retrieval methods. CER values are in regular scale. (a) True CER; (b) IPA retrieved CER; (c) UNet retrieved CER; (d) CAM retrieved CER. Highlighted regions are shown in bottom row.}
  \label{fig:resultscer}
  \vspace{-10pt}
\end{figure}

We present a visual comparison of various retrieval methods in Fig.~\ref{fig:resultscot} and Fig.~\ref{fig:resultscer} for COT and CER, respectively. When the sun is at an oblique angle, illuminating and shadow effects arise due to 3D radiative transfer~\cite{ademakinwa2023influence}. The highlighted regions show that IPA overestimates COT for regions with illuminating effects and underestimates for regions with shadowing effects. For CER, IPA shows the opposite trend, overestimating in shadowed regions and underestimating in illuminated areas. For COT retrievals, UNet struggles with thick cloudy regions (high COT areas). In contrast, our CAM model outperforms all other methods by effectively mitigating the impact of 3D radiative effects and accurately identifying COT in thick cloudy regions.

\section{Ablation Studies}\label{sec:ablation}
We conducted ablation studies to study the impacts of two key factors: the incorporation of attention mechanisms in UNet-style architectures and the choice of training loss function. Specifically, we systematically modified each factor independently while keeping the model architecture and training procedures otherwise identical. This approach allowed us to isolate their unique contributions and quantify their individual effects on model performance. The results are presented in Table~\ref{tab:ablation}.

\begin{table}[hbt]
    \centering
    \caption{Ablation Studies: Comparative Analysis on Attention Mechanism and objective function.}
    \label{tab:ablation}
    \resizebox{0.9\linewidth}{!}{%
    \begin{tabular}{r c c c}
        \toprule
        \textbf{Retrieval Methods} & 
        \textbf{Objective Function} & 
        \textbf{COT MAE ($\downarrow$)} & 
        \textbf{CER MAE ($\downarrow$)} \\
        \midrule
        UNet [w/o Attention] & L2 Loss & $0.065 \pm 0.008$ & $0.435 \pm 0.092$ \\
        CloudUNet & L2 Loss & $0.070 \pm 0.007$ & $0.407 \pm 0.049$ \\
        UNet [w/ Attention] & L2 Loss & $0.056 \pm 0.002$ & $0.425 \pm 0.097$ \\
        CAM [w/ Attention] & L2 Loss & $0.044 \pm 0.001$ & $0.291 \pm 0.011$ \\
        \midrule
        \textbf{CAM [w/ Attention]} & MTO Loss & $\mathbf{0.043} \pm \mathbf{0.001}$ & $\mathbf{0.252} \pm \mathbf{0.010}$ \\ 
        \bottomrule
    \end{tabular}
    }
\end{table}

\subsection{Impact of Attention Mechanisms}

\begin{figure}[hbt]
	\centering
	\includegraphics[width=.9\linewidth]{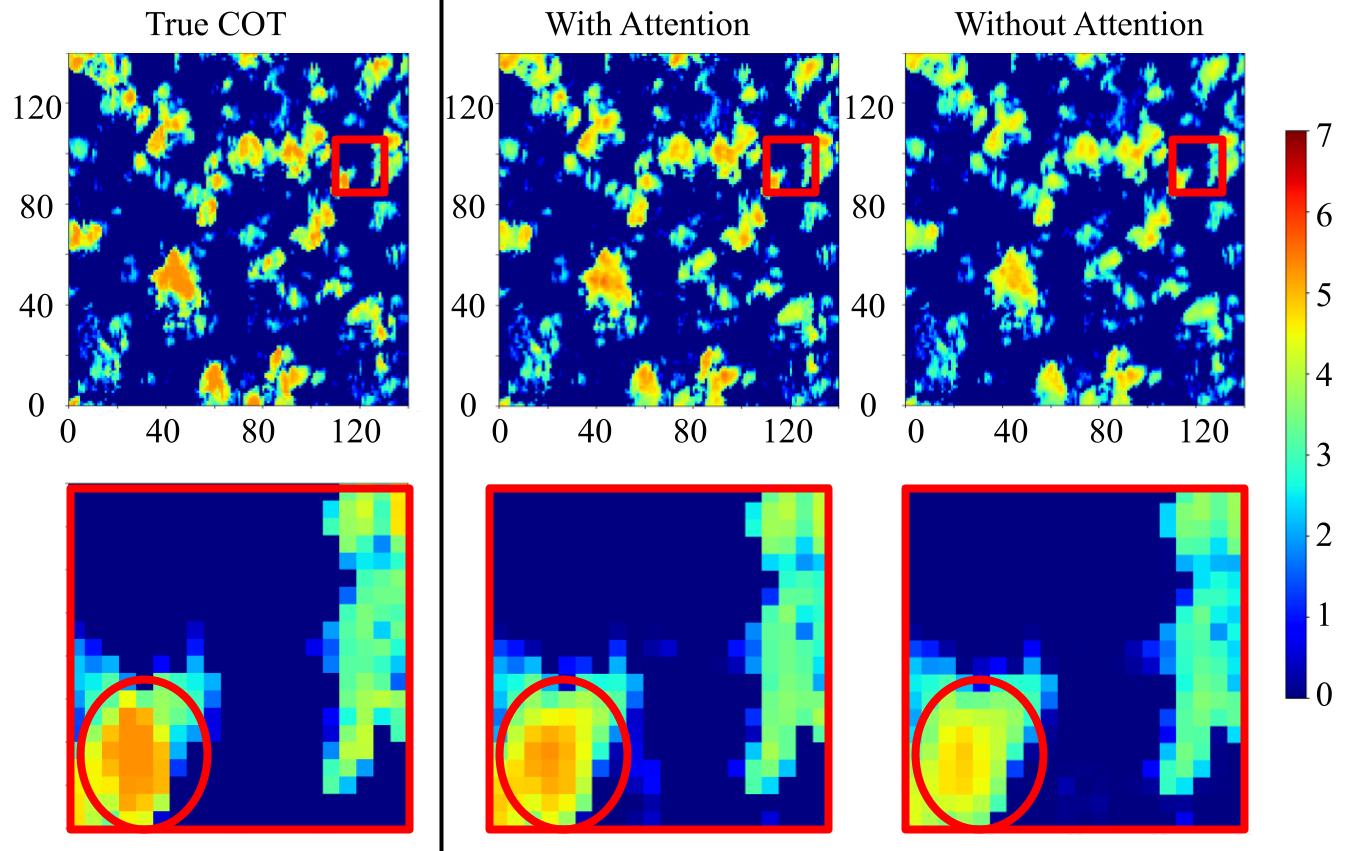}
	\caption{CAM COT retrieval with and without attention mechanism. Top row shows the COT profile while bottom row shows the highlighted regions. COT is shown in log scale.}\label{fig:attn}
\end{figure}

To assess the impact of attention mechanisms, we compare variants of UNet and our CAM model, with and without attention mechanisms. We notice that there is a considerable reduction in MAE when attention mechanism is used. This shows the generalizability of our proposed modification for UNet style architectures used in cloud property retrievals. The benefits of the attention mechanism are further illustrated in Fig.~\ref{fig:attn}, which shows CAM COT retrievals with and without attention mechanism. 

Without attention, COT retrievals in heterogeneous, thick cloud regions (high COT areas) were inaccurate due to the inherent tail distribution of COT values, which L2 loss fails to emphasize~\cite{tushar2024cloudunet}. However, with attention, the CAM model retrieved COTs in these regions more accurately.
However, with the attention mechanism, our CAM model retrieved COTs in these regions more accurately. This shows that the attention mechanism effectively captured rare but important features in the data, leading to more precise COT retrievals and reduced errors.

\subsection{Impact of Training Objective Function}
Different cloud properties have varying value ranges; for example, COT spans a much broader range than CER in typical atmospheric conditions. As a result, directly applying L2 loss during training may inadequately optimize CER retrieval. To address this imbalance, our multi-task objective (MTO) loss balances the L2 loss contributions for COT and CER, leading to a significant 13\% reduction in MAE for CER prediction. These results highlight the importance of normalizing training objectives to ensure effective estimation of cloud properties with differing dynamic ranges.

\section{Conclusions}
Retrieving cloud properties from radiance observations is a challenging problem due to 3D radiative transfer effects. In this paper, we proposed CAM model for jointly retrieving cloud optical thickness and cloud effective radius properties. Through quantitative and qualitative assessment using several evaluation metrics we demonstrated that multi-task objective function and attention mechanism are well suited for joint retrievals while minimizing the 3D radiative effects. In future, we plan to retrieve cloud properties from multi-angle radiance data where both solar zenith and view zenith angles vary.

\section*{Acknowledgments}
This research is partially supported by grants from NSF 2238743 and NASA 80NSSC21M0027.

This work was carried out using the computational facilities of the High Performance Computing Facility, University of Maryland Baltimore County. - https://hpcf.umbc.edu/

\small
\bibliographystyle{IEEEtranN}
\bibliography{ref}

\end{document}